\documentclass{article} 
\pdfoutput=1
\usepackage{times}
\usepackage{url}

\usepackage{floatrow}
\newfloatcommand{capbtabbox}{table}[][\FBwidth]
\usepackage{blindtext}
\usepackage{fullpage}
\usepackage{graphicx}
\usepackage{caption} 
\usepackage{subcaption} 
\usepackage{amsmath}

\title{Hidden Parameter Markov Decision Processes: A Semiparametric Regression Approach for Discovering Latent Task Parametrizations}

\author{
Finale Doshi-Velez\thanks{Both authors are primary authors on this occasion.}\\
Harvard Medical School \\
Boston, MA 02115 \\
\texttt{finale@alum.mit.edu} \\
\and
George Konidaris$^*$\\
Massachusetts Institute of Technology \\
Cambridge, MA 02139 \\
\texttt{gdk@csail.mit.edu}
}

\begin{document}

\maketitle

\begin{abstract}
Control applications often feature tasks with similar, but not
identical, dynamics.  We introduce the Hidden Parameter Markov
Decision Process (HiP-MDP), a framework that parametrizes a family of
related dynamical systems with a low-dimensional set of latent
factors, and introduce a semiparametric regression approach for
learning its structure from data.  In the control setting, we show
that a learned HiP-MDP rapidly identifies the dynamics of a new task
instance, allowing an agent to flexibly adapt to task variations.
\end{abstract}

\section{Introduction}
Many control applications involve repeated encounters with domains
that have similar, but not identical, dynamics.  An agent swinging a
bat may encounter several bats with different weights or lengths,
while an agent manipulating a cup may encounter cup with different
amounts of liquid.  An agent driving a car may encounter many
different cars, each with unique handling characteristics.

In all of these scenarios, it makes little sense of the agent to start
afresh when it encounters a new bat, a new cup, or a new car.
Exposure to a variety of related domains should correspond to faster
and more reliable adaptation to a new instance of the same type of
domain.  If an agent has already swung several bats, for example, we
would hope that it could easily learn to swing a new bat.  Why?  Like
many domains, the bat-swinging domain has a low-dimensional
representation that affects the system's dynamics in structured ways.
The agent's prior experience should allow it to both learn \emph{how}
to model related instances of a domain---such as the bat's length, a
latent parameter that smoothly changes in the bat's dynamics---and
\emph{what} specific model parameters (e.g., lengths) are likely.

Domains with closely-related dynamics are an interesting regime for
transfer learning. We introduce the Hidden Parameter
Markov Decision Process (HiP-MDP) as a formalization of these types of
domains, with two important features.  First, we posit that 
there exist a bounded number of latent parameters that, if known,
would fully specify the dynamics.  Second, we assume the parameter
values remain fixed for a task's duration (e.g. the bat's length will
not change during a swing), and the agent will know when a change has
occurred (getting a new bat).

The HiP-MDP parameters encode the minimum amount of learning required
for the agent to adapt to a new domain instance.  Given a generative
model of how the latent parameters affect domain dynamics, an agent could
rapidly identify the dynamics of a particular domain instance by
maintaining and updating its distribution (or \textit{belief}) over
the latent parameters.  Instead of learning a new policy for each
domain instance, it could synthesize a parametrized control policy
\cite{Kober12,daSilva12} based on a point estimate of the parameter
values, or plan in the belief space over its parameters
\cite{beetle,pomcp,contba,guez12,learning-to-plan}.

We present a method for learning the structure of a HiP-MDP
from data. Our generative model uses Indian Buffet Processes \cite{ibp}
to model what latent parameters are relevant for a particular set of
dynamics and Gaussian processes \cite{gp} to model the dynamics
functions.  We do not require knowledge of a system's kinematics
equations, nor must we specify the number of latent parameters in
advance.  Our HiP-MDP model efficiently performs control in the
challenging acrobot domain \cite{Sutton98} by rapidly identifying the
dynamics of new instances.

\vspace{-.1in}

\section{Background}
\label{sec:background}
\paragraph{Bayesian Reinforcement Learning}
The reinforcement learning setting consists of a series of
interactions between an agent and an environment.  From some state
$s$, the agent chooses an action $a$ which transitions it to a new
state $s'$ and provides reward $r$.  Its goal is to maximize its
longterm expected rewards, $E[\sum_t \gamma^t r_t]$, where $\gamma \in
[0,1)$ is a discount factor that weighs the relative importance of
  near-term and long-term rewards.  This series of interactions can be
  modeled as a Markov Decision Process (MDP), a 5-tuple
  $\{S,A,T,R,\gamma\}$ where $S$ and $A$ are sets of states $s$ and
  actions $a$, the transition function $T(s'|s,a)$ gives the
  probability of the next state being $s'$ after performing action $a$
  in state $s$, and the reward function $R(s,a)$ gives the reward $r$
  for performing action $a$ in state $s$.  We refer to the transition
  function $T(s'|s,a)$ as the \emph{dynamics} of a system.

The transition function $T$ or the reward function $R$ must be learned
from experience.  Bayesian approaches to reinforcement learning
\cite{beetle, contba, pomcp} place a prior over the transition
function $T$ (and sometimes also $R$), and refine this prior with
experience.  Thus, the problem of learning an unknown MDP is
transformed into a problem of planning in a known
\emph{partially}-observable Markov Decision Process (POMDP).  A POMDP
\cite{Kaelbling98} consists of a 7-tuple
$\{Y,A,O,\tau,\Omega,R,\gamma\}$, where $Y, A,$ and $O$ are sets of
states $y$, actions $a$, and observations $o$; $\tau(y'|y,a)$ and
$R(y,a)$ are the transition and reward functions; and the observation
function $\Omega(o|y',a)$ is the probability of receiving an
observation $o$ when taking action $a$ to state $y'$.  Bayesian RL
learns an MDP by planning in a POMDP with states $y_t = \{s_t,T\}$:
the fully-observed ``world-state'' $s_t$ and the hidden dynamics $T$.
 
However, solving POMDPs in high-dimensional,
continuous state spaces remains challenging, 
despite advances in POMDP planning
\cite{pomcp, sarsop,guez12}, including situations with
mixed-observability \cite{momdp}.  The HiP-MDP, which we
introduce in section~\ref{sec:hipmdp}, simplifies the Bayesian RL
challenge by using instances of related tasks to first find a
low-dimensional representation of the transition function $T$.

\paragraph{Indian Buffet Processes and Gaussian Processes}
Our specific instantiation of the HiP-MDP uses two models from
Bayesian nonparametric statistics.  The first is the Indian Buffet
Process (IBP).  The IBP is a prior on 0-1 matrices $M(n,k)$ with a
potentially unbounded number of columns $k$.  To generate samples from
the prior, we first use a Beta process to assign a probability $p_k$
to each column $k$ such that $\sum_k p_k$ is bounded.  Then, each
entry $M(n,k)$ is set to 1 independently with probability $p_k$.  The
IBP has the property that the distribution of the number of nonzero
columns for any row is $Pois(\alpha)$, but some columns will be very
popular (that is, a few $p_k$ will be large), while others will be
used in only a few rows.  We will use the IBP as a prior on what
latent parameters are relevant for predicting a certain transition
output.

The second model we use is the Gaussian Process (GP).  A GP is a prior
over continuous functions $y=f(x)$ where the prior probability of a
set of outputs $\{y_1,...,y_t\}$ given a set of inputs
$\{x_1,...,x_t\}$ is given by a multivariate Gaussian $N(m,K)$, where
$m(x_i)$ is the mean function of the Gaussian process and the
covariance matrix has elements $K(x_i,x_j)$ for some positive definite
kernel function $K$.  We use additive mixtures of Gaussian processes
to model the transition function.

\vspace{-.1in}

\section{Hidden Parameter Markov Decision Processes}
\label{sec:hipmdp}
We focus on learning the dynamics $T$ and assume that the reward
function $r =R(s,a)$ is fixed across all instances (e.g., the agent
always wants to swing the bat).  Let $b$ denote each instance of a
domain.  The Hidden Parameter Markov Decision Process (HiP-MDP) posits
that the variation in the dynamics of a different instances can be
captured through a set of hidden parameters $\theta_b$.

The HiP-MDP is be described by a tuple: $\{S, A, \Theta, T, R, \gamma,
P_\Theta\}$, where $S$ and $A$ are the sets of states $s$ and actions
$a$, and $R(s,a)$ is the reward function.  The dynamics $T$ for each
instance $b$ depends on the value of these hidden parameters: 
$T(s'|s,a,\theta_b)$.  We denote the set of all possible parameters
$\theta_b$ with $\Theta$ and let $P_\Theta$ be the prior over these
parameters.  Thus, a HiP-MDP describes a \textit{class} of tasks; a
particular instance of that class is obtained by fixing the parameter
vector $\theta_b \in \Theta$.  Specifically, we assume that $\theta_b
\sim P_\Theta$ is drawn at the beginning of each task instance $b$ and
does not change until the beginning of the next instance.

In special cases, we may be able to derive analytic expressions for
how $\theta_b$ affects the dynamics $T( s' | s , \theta_b , a )$: for
example, in a manipulation domain we might be able to derive the
kinematic equations of how the cup will respond to a force given a
certain volume of liquid.  However, in most situations, the
simplifications required to derive these analytical forms for the
dynamics will be brittle at best.  The IBP-GP prior for the HiP-MDP,
presented in section~\ref{sec:ibp-gp}, describes a semiparametric
approach for modeling the dynamics that places few assumptions on the
form of the transition function $T( s' | s , \theta_b , a )$ while
still maintaining computational tractability.

As pointed out by Bai et al. \cite{learning-to-plan} in a similar
setting, we can consider the HiP-MDP  a type of POMDP where the
hidden state are the parameters $\theta_b$.  However, a HiP-MDP makes
two assumptions which are stronger than those of a POMDP.  First, each
\emph{instance} of a HiP-MDP is an MDP---conditioned on $\theta_b$ the
transition and reward functions obey the Markov property. Thus, we
could always learn to solve each HiP-MDP instance as its own distinct
MDP. Second, the parameter vector $\theta_b$ is fixed for the duration
of the task, and thus the hidden state has no dynamics. This
considerably simplifies the procedure for inferring the hidden
parametrization.

\vspace{-.05in}

\section{The IBP-GP HiP-MDP}
\label{sec:ibp-gp}
Let the state $s$ be some $d$-dimensional vector of continuous,
real-valued variables.  We propose a transition model $T$ of the form:
\begin{eqnarray*}
  (s_d' - s_d ) &\sim& \sum^K_k z_{kad} w_{kb} f_{kad}( s ) + \epsilon
  \\ \epsilon &\sim& N( 0 , \sigma_{nad}^2 ).
\end{eqnarray*}
The weights $w_{kb}$ are the values associated with the latent
parameters for instance $b$.  The filter parameters $z_{kad} \in \{0,1\}$
denote whether the $k^{th}$ latent parameter is relevant for making
predictions about dimension $d$ when taking action $a$.  The
task-specific basis functions $f_{kad}$ describe how a change in the
latent factor $w_{kb}$ affects the dynamics.  The additivity
assumption in our semi-parametric basis function regression allows us
to learn all the latent elements of this model: the number of factors
$K$, the weights $w_{kb}$, the filter parameters $z_{kad}$, and the
form of the basis functions $f_{kad}$.

\paragraph{Generative Model}
The IBP-GP Hip-MDP places the following priors on the three sets of
hidden variables, $w_{kb}, z_{kad}$, and $f_{kad}$:
\begin{eqnarray*}
  z_{kad} &\sim& IBP( \alpha ) \textrm{ for } k > 1\\ 
  f_{kad} &\sim& GP( \psi ) \\
  \mu_{w_k} &\sim& N(  0 , \sigma_{w_0}^2 ) \\
  w_{kb} &\sim& N( \mu_{w_k} , \sigma_w^2 ) \textrm{ for } k > 1\\
  w_{1b}, z_{1ad} &=& 1,  
\end{eqnarray*}
where $\alpha$ and $\psi$ are the parameters of the IBP and GP.
Fixing $z_{1ad} = 1$ and $w_{1b} = 1$ sets the scale for latent
parameters and makes the first basis function $f_{1ad}$ the mean dynamics
of the domain.

The IBP-GP prior encodes the assumption that for any domain instance
$b$ the real world contains a countably infinite number of independent
scalar latent parameters $w_{kb}$.  However, when making predictions
about a specific state dimension $d$ given action $a$, only a few of
these infinite possible latent parameters will be relevant.  Thus, we
only need to infer the values of a finite number of weights $w_{kb}$
to characterize the dynamics of a finite number of state dimensions
under a finite set of actions.  Using an IBP prior on the filter
parameters $z_{kad}$ implies that we expect a few latent factors to be
relevant for making most of predictions.  Moreover, additional
prediction tasks---such as a new action, or a new dimension to the
state space---can be incorporated consistently.  As a regression
model, IBP-GP model is an infinite version of the Semiparametric
Latent Factor Model \cite{teh_semi}.

\paragraph{Batch Inference}
We focus on scenarios in which the agent is given a large amount of
\emph{batch} observational data from several domain instances and
tasked with quickly performing well on new instances.  Our batch inference
procedure uses the observational data to fit the filter parameters
$z_{kad}$ and basis functions $f_{kad}$, which are independent of any
particular instance, and compute a posterior over the weights
$w_{kb}$, which depend on each instance.  These settings of $z_{kad},
f_{kad}$, and $P(w_{kb})$ will used to infer the instance-specific
weights $w_{kb}$ efficiently online.

The posterior over the weights $w_{kb}$ is Gaussian given the filter
settings $z_{kad}$ and means $\mu_{w_k}$.  However, marginalization
over the basis functions $f_{kad}$ requires computing inverses
of matrices of size $N = \sum_b n_b$, where $n_b$ is the number of
data collected in instance $b$.  Instead, we represented each function
$f_{kad}$ by a set of $( s^*, f_{kad}(s^*) )$ pairs for states $s^*$
in a set of support points $S^*$.  Various optimization procedures
exist for choosing the support points~\cite{gp-pseudoinput, teh_semi};
we found that iteratively choosing support points from existing points
to minimize the maximum reconstruction error within each batch was
best for a setting in which a few large errors can result in poor
performance.

Given a set of tuples $(s,a,s',r)$ from a task instance $b$, we first
created tuples $( s^* , a , \Delta_b(s^*) )$ for all $s^* \in S^*$ and all
actions $a$, all dimensions $d$, and all instances $b$.  The tuple $(
s_d* , a , \Delta_b(s_d*) )$ can be predicted based on all other data
available for that action $a$, dimension $d$ pair using standard
Gaussian process prediction:
\begin{equation*}
E[ \Delta(s_d*) ] = K_{s^*S_{ab}} (K_{S_{ab} S_{ab}} + \sigma_{nad}^2 I)^{-1} \Delta_b(S_{abd}),
\end{equation*}
where $S_{ab}$ is the collection of tuples $(s,a,s',r)$ with action
$a$ from instance $b$, $K_{s^*S_{ab}}$ is the vector $K(s^*,s)$ for
every $s \in S_{ab}$, $K_{S_{ab} S_{ab}}$ is the matrix $K(s,s)$, and
$\Delta_b(S_{abd})$ is the vector of differences $s_d' - s_d$. % for .
This procedure was repeated for every task instance $b$. 

Following this preprocessing step, we proceeded to infer the filter
parameters $z_{kad}$, the weights $w_{kb}$, and the values of the
task-specific basis functions at the support points $f_{kad}(S^*)$
using a blocked Gibbs sampler.  Given $z_{kad}$ and $w_{kb}$, the
posterior over $f_{kad}(S^*)$ is Gaussian.  Let $f_{ad}(S^*)$ be a
column vector of concatenated $f_{kad}(S^*)$ vectors, and let
$\Delta(S^*)$ be a column vector of concatenated $\Delta_b(S^*)$
vectors.  The mean and covariance of $f_{ad}(S^*)$ is given by
\begin{eqnarray*}
\textrm{cov}( f_{ad}(S^*)) &=& \sigma_{nad}^2 ( W^{T}W + \sigma_{nad}^2 {\bf K}_{S^*S^*}^{-1} )^{-1} \\ 
E[ f_{ad}(S^*) ] &=& \frac{1}{\sigma_{nad}^2} \textrm{cov}( f_{ad}(S^*))^{-1} W^{T} * \Delta(S^*)
\end{eqnarray*}
where 
\begin{eqnarray*}
{\bf K } &=& I_k \otimes K_{S^*S^*}  \\
W &=& w_{b}(z_{kad}) \otimes I_{|S^*|}, 
\end{eqnarray*}
where $\otimes$ is the Kronecker product, $K_{S^*S^*}$ is the matrix
$K(s^*,s^*)$ for every $s^* \in S^*$, and we write $w_{b}(z_{kad})$ to be
$B$ by $k'$ matrix of elements $w_{kb}$ such that $z_{kad} = 1$.  We
repeat this process for each action $a$ and each dimension $d$.

Similarly, given $z_{kad}$ and $f_{kad}(S^*)$, the posterior over
$w_{kb}$ is also Gaussian.  For each instance $b$, the mean and
covariance of $w_{b}(z_{kad})$ (that is, the weights not set to zero) are given by:
\begin{eqnarray*}
  \textrm{cov}( w_{kb} ) &=&  \sigma_n^2  ( F_{b}^{T} F_{b} + I_{k-1} \frac{ \sigma_n^2 }{ \sigma_w^2 } )^{-1} \\
  E[ w_{kb} ] &=& \textrm{cov}( w_{kb} )^{-1}  ( \frac{ \mu_{w_k} }{ \sigma_w^2 } + \frac{1}{\sigma_n^2} F_{b}^{T} \\
 & & \cdot ( \Delta_b(S^*) - F_{1b}(S^*) ),
\end{eqnarray*}
for $k > 1$, where $F_{b}$ is matrix with $k'-1$ columns concatenating
values for $f_{kad}$ for all actions $a$ and all dimensions $d$ and
excluding $k = 1$, and $\Delta_b(S^*)$ is a column vector concatenating
the differences for all actions $a$ and all dimensions $d$.

To sample $z_{kad}$ for an already-initialized feature $k$, we note
that the likelihood of the model given $z$, $w$, and $f$ with some
$z_{k'ad} = 0$ is Gaussian with mean and variance
\begin{eqnarray}
  E[ \Delta{s^*} ] &=& \sum_{k \neq k'} z_{kad} w_{kb} f_{kad}(s^*) \\ \label{eqn:marg_ll_mu}
  \textrm{cov}( \Delta{s^*} ) &=& \sigma_{nad}^2 \label{eqn:marg_ll_cov0}.
\end{eqnarray}
If $z_{kad} = 1$, then likelihood is again Gaussian with the same mean
(here we assume that the GP prior on $f$ is zero-mean) but with covariance
\begin{equation}
  \textrm{cov}( \Delta{s^*} ) = w_{k'}w_{k'}^{T} \otimes K_{S^*S^*} + \sigma_{nad}^2 I,
  \label{eqn:marg_ll_cov1}
\end{equation}
where $w_{k'}$ is a vector of latent parameter values for all
instances $b$.  We combine these likelihoods with the prior from
section~\ref{sec:background} to sample $z_{kad}$.

Initializing a new latent parameter $k'$---that is, a $k'$ for which
we do not already have values for the weights $w_{k'b}$---involves
computing the marginal likelihood $P(\Delta(S^*)|z_{k'ad}, w_{kb},
f_{kad})$, which is intractable.  We approximate the likelihood by
sampling $N_w$ sets of new weights $w_{k'b}$.  Given values for the
weights $w_{k'b}$ for each instance $b$, we can compute the likelihood
with the new basis function $f_{k'ad}$ marginalized out using
equations~\ref{eqn:marg_ll_mu} and~\ref{eqn:marg_ll_cov1}.  We average
these likelihoods to estimate the marginal likelihood of $z_{k'ad} =
1$ for the new $k'$.  (For the $z_{k'ad} = 0$ case, we can just use
the variance from equation~\ref{eqn:marg_ll_cov0}.)  If we do set
$z_{k'ad} = 1$ for the new $k'$, then samples a set of weights
$w_{k'b}$ from our $N_w$ samples based on their importance weight
(marginal likelihood).

Finally, given the values of the weights $w_{kb}$ for a set of
instances $b$, we can update the posterior over $\mu_{w_k}$ with a
standard conjugate Gaussian update:
\begin{eqnarray*}
\textrm{var}( \mu_{w_k} ) &=& (\frac{1}{\sigma_{w_0}^2} + \frac{B}{\sigma_w^2})^{-1} \\
E[ \mu_{w_k} ] &=& \frac{ \sum_b w_{kb} }{ \sigma_w^2 } \textrm{var}( \mu_{w_k} )^{-1}.
\end{eqnarray*}
We use this posterior over the weight means $\mu_{w_k}$ when we
encounter a new instance $b'$.

\paragraph{Online Filtering}
Given values for the filter parameters $z_{kad}$ and the basis
functions $f_{kad}$, the posterior on the weights $w_{kb}$ is
Gaussian.  Thus, we can write the parametrized belief $b_t(w_{kb})$ at
some time $t$ with $( h_w, P_w )$, where $P_w$ is the inverse
covariance $\Sigma_w^{-1}$ and $h_w$ is the information mean $\mu_w
P_w$.  Then the update given an experience tuple $(s,a,s',r)$ is given
by
\begin{eqnarray*}
h_w(t+1) &=& h_w(t) + F_{a}^{T} \Sigma_{na}^{-1} \Delta(s) \\
P_w(t+1) &=& P_w(t) + F_{a}^{T} \Sigma_{na}^{-1}  F_{a}, 
\end{eqnarray*}
where $F_{a}$ is a $d$ by $k$ matrix of basis values $f_{kad}(s)$,
$\Sigma_{na}$ is a $d$ by $d$ noise matrix with $\sigma_{nad}^2$ on
the diagonal (note that therefore the inverse $\Sigma_{na}^{-1}$ is
trivially computed), and $\Delta(s)$ is a $d$-dimensional vector of
$s'_d - s_d$.  If updates are performed only every $n$ time steps, we
can simply extend $F$, $\Sigma$, and $\Delta(s)$ to be $nd$ by $k$,
$nd$ by $nd$, and $nd$ respectively.

Computing $F_{a}$ requires computing the values of the basis functions
at the point $s$.  Since we only have the values computed at
pseudo-input points $s^* \in S^*$, we use our standard GP prediction
equation to interpolate the value for this new point:
\begin{equation*}
E[ f_{kad}(s) ] = K_{sS^*} (K_{S^*S^*} + \sigma_{nad}^2 I)^{-1} f_{kad}(S^*),
\end{equation*}
where $K_{sS^*}$ is the vector $K(s,s^*)$ for every $s^* \in S^*$,
$K_{S^*S^*}$ is the matrix $K(s^*,s^*)$, and $f_{kad}(S^*)$ is the
vector $f_{kad}(s^*)$.  Using only the mean value $f_{kad}(s^*)$
ignores the uncertainty in the basis function $f_{kad}$.  While
incorporating this variance is mathematically straight-forward---all
updates remain Gaussian---it adds additional computations to the
online calculation.  We found that using only the means already
provided significant gains in learning in practice.

\section{Results}
In this section, we describe results on two benchmark problems:
cartpole and acrobot \cite{Sutton98}.  In all of our
tests, we use an anisotropic squared-exponential kernel with length
and scale parameters approximated for each action $a$ and dimension
$d$.

\subsection{Cartpole}

The cartpole task begins with a pole that is initially standing
vertically on top of a cart.  The agent may apply a force either to
the left or the right of the cart to keep the pole from falling over.
The domain has a four-dimensional state space $s = \{ x , \dot{x} ,
\theta , \dot{\theta}\}$ consisting of the cart's position $x$, the
cart's velocity $\dot{x}$, the pole's angle $\theta$, and the pole's
angular velocity $\dot{\theta}$.  At each time step of length $\tau$,
the system evolves according to the following equations:
\begin{eqnarray}
x_{t+1} =& x_{t} + \tau \dot{x}_t\\ \nonumber
\dot{x}_{t+1} =&\dot{x}_t + \tau(v  - ml \ddot{\theta}\cos{\theta} / M)\\ \nonumber
\theta_{t+1} =& \theta_t + \tau \dot{\theta}_t\\ \nonumber
\dot{\theta}_{t+1} =& \dot{\theta}_t + \tau \ddot{\theta}, 
\label{eqn:cartpole_dynamics}
\end{eqnarray}
where $v = \frac{f + ml \dot{\theta}_t^2 \sin{\theta}}{M},$ $
\ddot{\theta} = \frac{(g \sin{\theta} - v\cos{\theta})}{ (l
  (\frac{4}{3} - m\cos{\theta}^2/M)}$, $f$ is the applied force, $g$
is gravity, $M$ is the mass of both the cart and the pole, and $m$ and
$l$ are the mass and length of the pole, respectively.

\begin{table*}[t]
\centering
\caption{Mean-Squared Error on Cartpole (with 95\% confidence intervals)} 
\begin{tabular}{|p{.3in}|p{1in}|p{1in}|p{1in}|p{1.2in}|} \hline
 & Batch + Train  & Batch Only  & Train Only  & IBP-GP HIP-MDP \\ \hline 
$\dot{x}$ & 4.8e-06 (1.7e-08)  & 4.8e-06 (1.7e-08)  & 9.1e-05 (2.3e-07)  & 4.7e-06 (1.7e-08) \\ \hline 
$\ddot{x}$ & 1.1e-07 (1.0e-08)  & 1.1e-07 (1.1e-08)  & 7.6e-07 (1.4e-07)  & 3.3e-08 (2.9e-09) \\ \hline 
$\dot{\theta}$ & 1.1e-03 (3.2e-06)  & 1.1e-03 (3.2e-06)  & 3.4e-02 (1.4e-04)  & 1.0e-03 (3.2e-06) \\ \hline 
$\ddot{\theta}$ & 1.5e-05 (1.5e-06)  & 1.5e-05 (1.6e-06)  & 3.7e-05 (2.8e-06)  & 2.5e-06 (1.1e-07) \\ \hline 
All & 2.7e-04 (8.9e-04)  & 2.7e-04 (8.9e-04)  & 8.4e-03 (2.8e-02)  & 2.6e-04 (8.8e-04) \\ \hline 

\end{tabular}
\label{tab:cartpole_mse}
\end{table*}

We varied the pole mass $m$ and the pole length $l$.  Cartpole is a
simple enough domain such that changing these parameters does not
change the optimal policy---if the pole is falling to the left, the
cart should be moved left; if the pole is falling to the right, the
cart should be moved right.  However, we used cartpole to demonstrate
the quality of predictions and describe the latent parameters.

For each training $(m,l)$ setting, the agent received batches of data
from Sarsa \cite{Sutton98} (using a $3$rd order Fourier Basis
\cite{fourier}) run for five repetitions of 30 episodes, where each
episode was run for 300 iterations or until the pole fell down.  Data
from seven $(m, l)$ settings $\{$ (.1,.4) , (.3,.4) , (.15,.45) ,
(.2,.55) , (.25,.5) , (.3,.4) , (.3,.6) $\}$ were reduced to 750
support points and then the batch inference procedure
(section~\ref{sec:hipmdp}) was repeated 5 times for 250 iterations
with $\sigma_w = 4$, $\alpha = 2$ and the Gaussian process
hyper-parameters set from the first batch.

Next, 50 training points were selected from Sarsa run on all $( m ,
l)$ settings with $m \in \{ .1 , .15 , .2 , .25 , .3 \}$ and $l \in \{
.4 , .45 , .5 , .55 , .6 \}$.  The online inference procedure
(section~\ref{sec:hipmdp}) was used to estimate the weights $w_{kb}$
given the filter parameters $z_{kad}$ and basis functions $f_{kad}$
from the batch procedure.  The quality of the predictions $\Delta(s)$
was evaluated on 50 (different) test points.  Our HiP-MDP predictions
were compared to using all of the batch data in a single GP (ignoring
the fact that the data came from different instances), using only the
50 training points from the current instance, and combining both the
50 training points and batch data together in one GP.

\begin{figure*}[tb]

\begin{subfigure}{.45\textwidth}
\includegraphics[width=\textwidth]{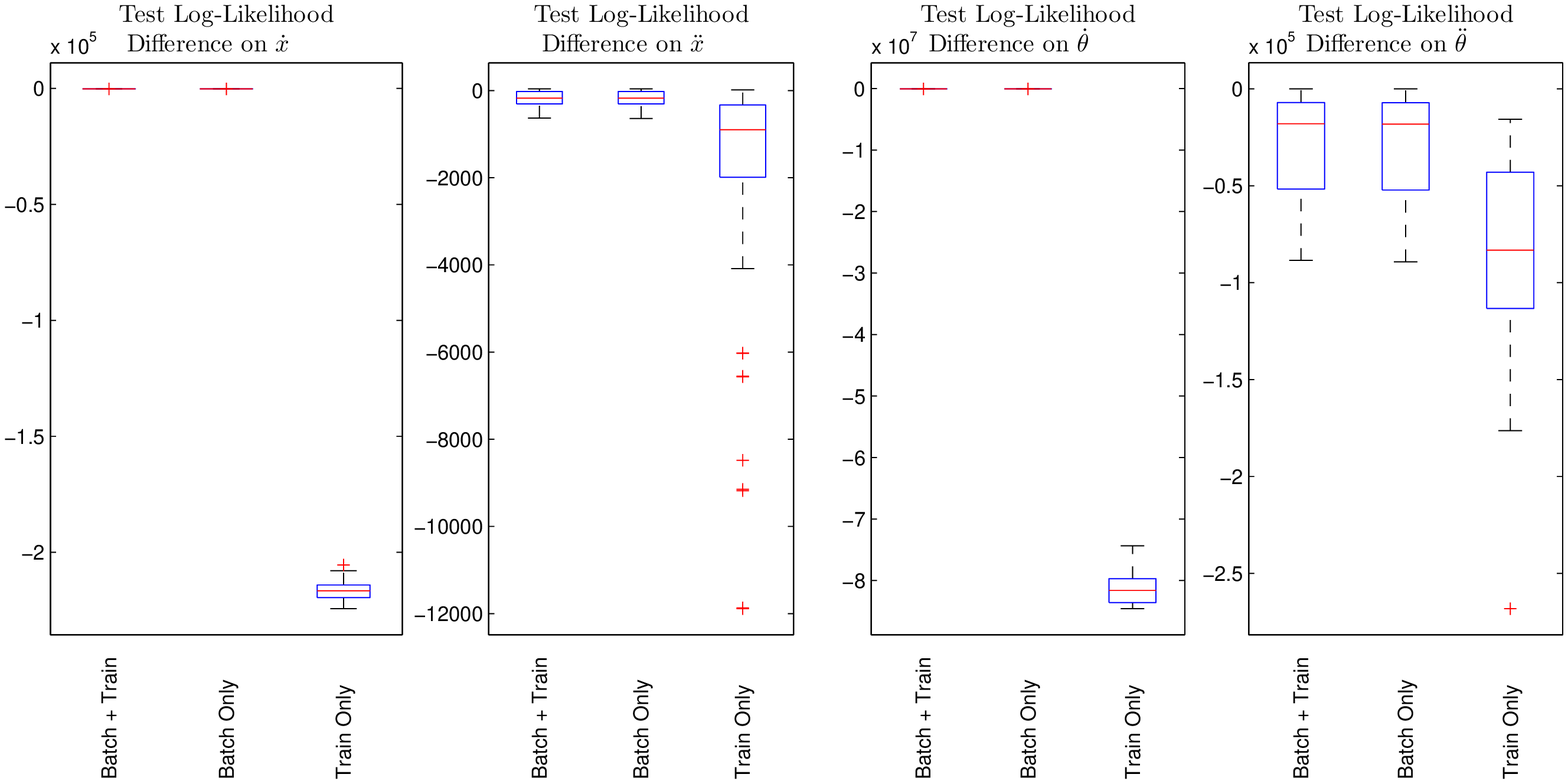}
\caption{Difference in Test Log-Likelihood on Cartpole.  Negative
  values indicate performance worse than the IBP-GP}
\label{fig:cartpole_ll}
\end{subfigure}
~
\begin{subfigure}{.45\textwidth}
\includegraphics[width=\textwidth]{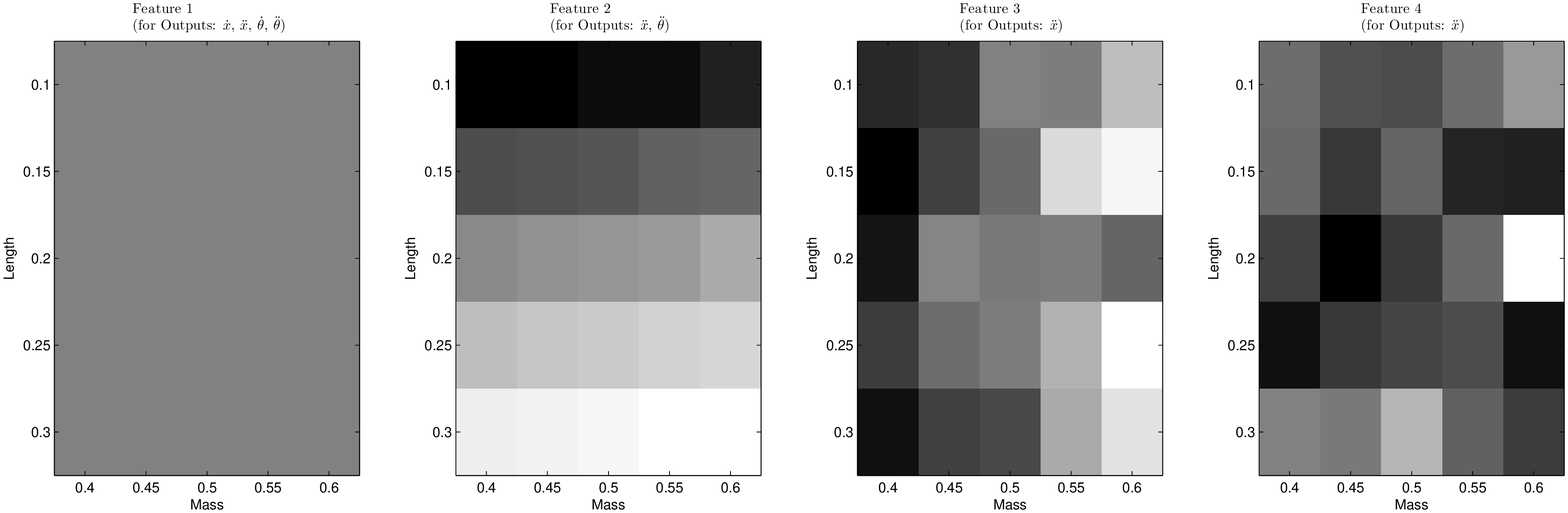}
\caption{Cartpole Weights vs. Mass and Length}
\label{fig:cartpole_m12}
\end{subfigure}
~
\begin{subfigure}{.45\textwidth}
\includegraphics[width=\textwidth]{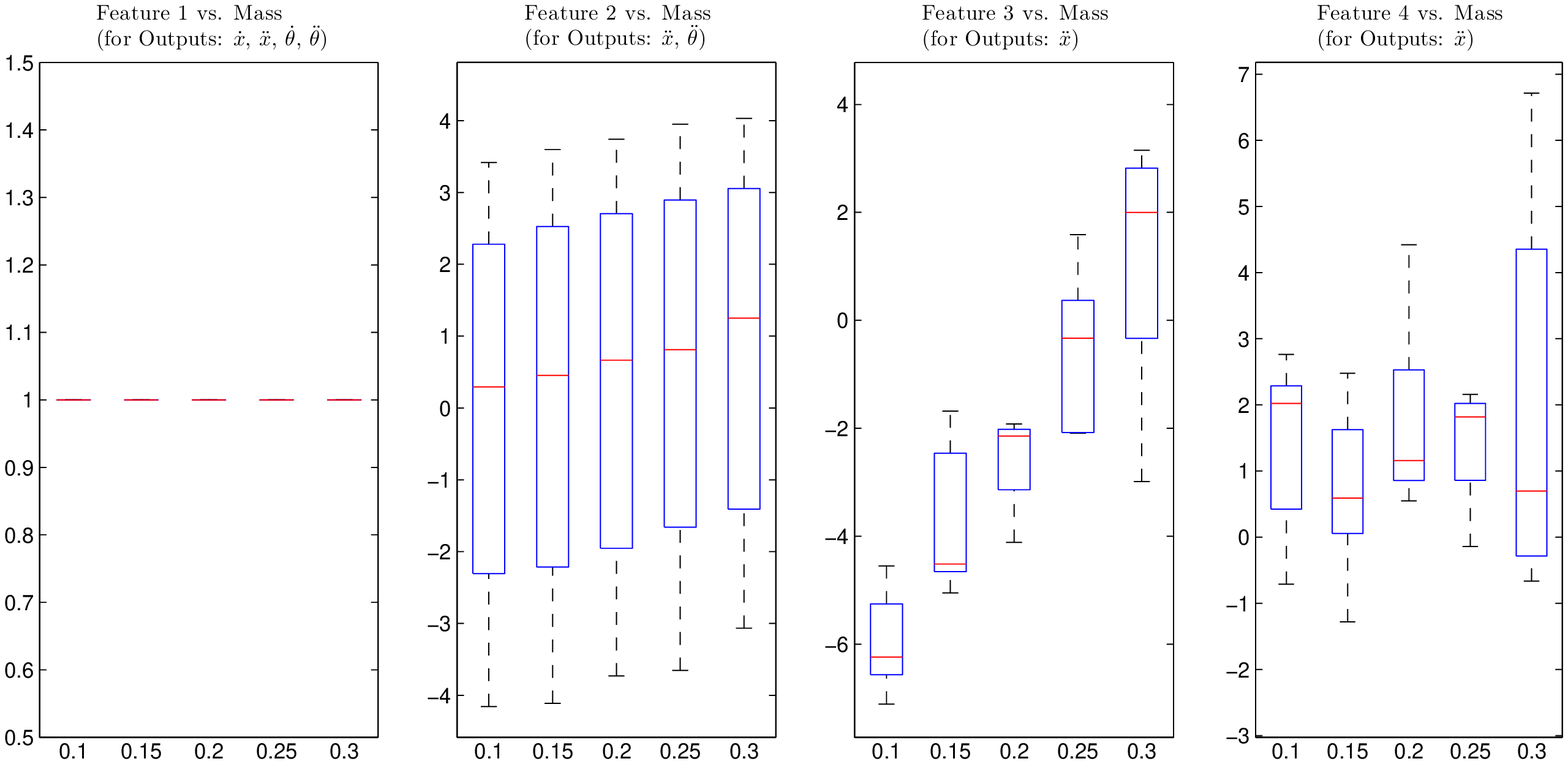}
\caption{Cartpole Weights vs. Mass}
\label{fig:cartpole_m1}
\end{subfigure}
~
\begin{subfigure}{.45\textwidth}
\includegraphics[width=\textwidth]{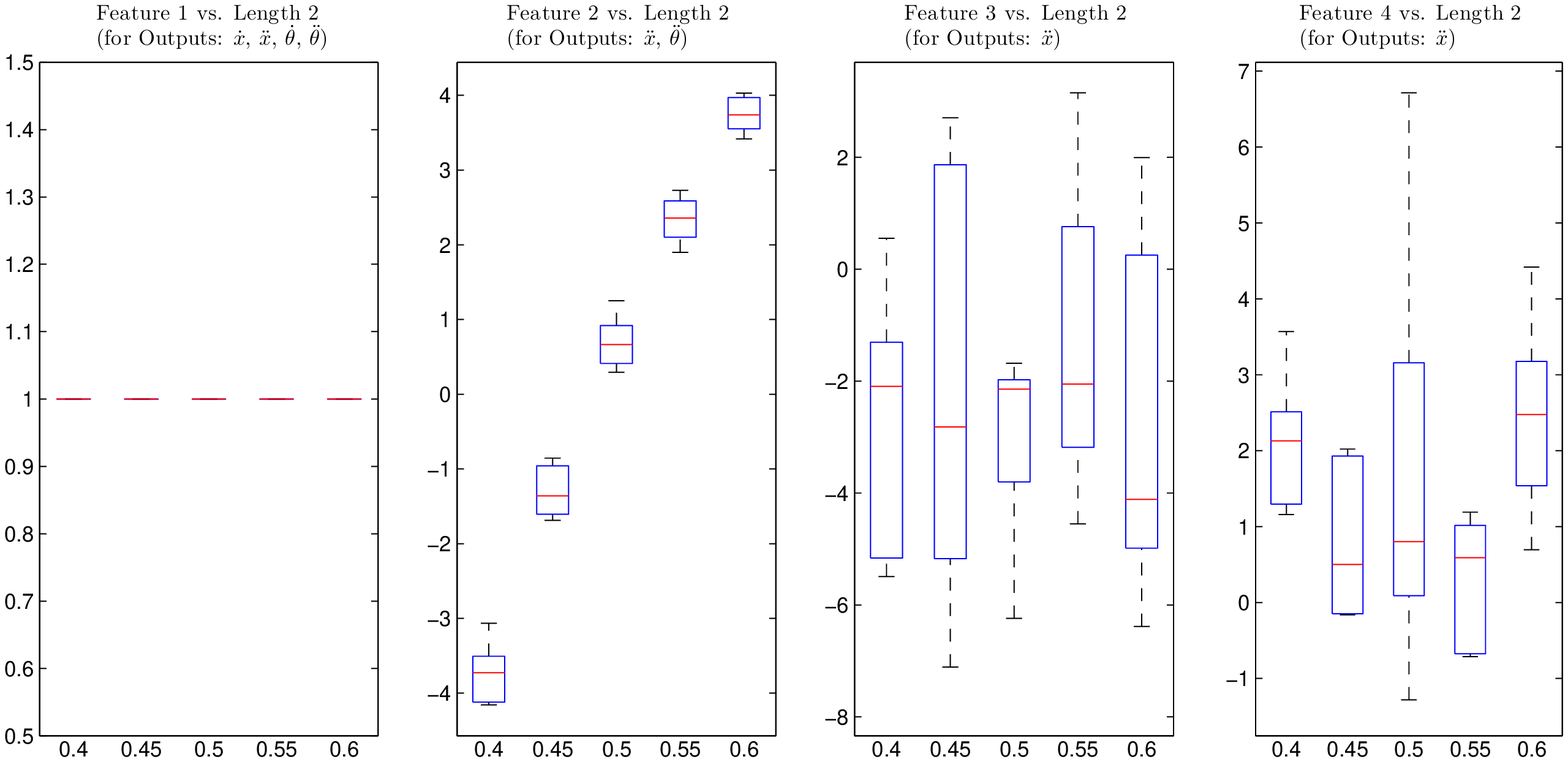}
\caption{Cartpole Weights vs. Length}
\label{fig:cartpole_m2}
\end{subfigure}
\caption{Experimental Results for Cartpole.}
\end{figure*}

The relative test log-likelihoods are summarized in
figure~\ref{fig:cartpole_ll}, and table~\ref{tab:cartpole_mse}
summarizes the corresponding mean-squared errors across the 5 runs.
Using only the training points from that instance has the highest
error across all dimensions $x$, $\dot{x}$, $\theta$, and
$\dot{\theta}$.  Our IBP-GP approach performs similarly to just
applying a single GP to the batch data for predicting the change in
outputs $x$ and $\theta$; these two dimensions depend only on previous
values of $\dot{x}$ and $\dot{\theta}$ and not on the pole mass $m$ or
length $l$ (see equations~\ref{eqn:cartpole_dynamics}).  Our approach
significantly outperforms all baselines for outputs $\dot{x}$ and
$\dot{\theta}$, whose changes do depend on the parameters of system.

In all 5 of the MCMC runs, a total of 4 latent parameters were
inferred.  The output dimensions $x$ and $\theta$ consistently only
used the first (baseline) feature---that is, our IBP-GP's predictions
were in fact the same as using a single GP.  This observation is
consistent with the cartpole dynamics and the observed prediction
errors in figure~\ref{fig:cartpole_ll} and
table~\ref{tab:cartpole_mse}.  The second feature was used by both
$\dot{x}$ and $\dot{\theta}$ and was positively correlated with both
the pole mass $m$ and the pole length $l$
(figures~\ref{fig:cartpole_m1},~\ref{fig:cartpole_m12},
and~\ref{fig:cartpole_m12}).  In
equations~\ref{eqn:cartpole_dynamics}, both $\dot{x}$ and
$\dot{\theta}$ have many $ml$ terms.  The third consistently
discovered feature was used only by $\dot{x}$ and was positively
correlated with the pole mass $m$ and not correlated with the pole
length $l$; the equations for $\dot{x}$ have several terms that depend
only on $m$.  The fourth feature (used only to predict $\dot{x}$) had
the highest variability; it generally has higher values for more
extreme length settings, suggesting that it might be a correction for
more complex nonlinear effects.

\subsection{Acrobot}

The acrobot domain features a double-pendulum.  The agent can apply a
positive, negative, or neutral torque to the joint between the two
poles, and its goal is to swing the pendulum up to vertical.  The
four-dimensional state space consists of the angle $\theta_1$ and
angular velocity $\dot{\theta_1}$ of the first segment (with mass
$m_1$) and the angle $\theta_2$ and angular velocity $\dot{\theta_2}$
of the second segment (with mass $m_2$).  

For batch training, we used mass settings $(m_1,m_2)$ settings of $\{$
(.7,.7) , (.7,1.3) , (.9,.7) , (.9,1.1) , (1.1,.9) , (1.1,1.3) ,
(1.3,.7) , (1.3, 1.3) $\}$, again employing Sarsa (this time with a $5$th order Fourier Basis). 
 First, 1000 support points (and GP
hyper-parameters) were chosen to minimize the maximum prediction error
within each batch.  Next, we used the approach in
Section~\ref{sec:hipmdp} to infer the filter parameters $z_{kad}$
and get approximate MAP-estimates for the basis functions $f_{kad}$.

\begin{figure*}[tb]
\CenterFloatBoxes
\begin{floatrow}
\ffigbox{
\centering
\vspace{-12pt}
\includegraphics[width=2.7in]{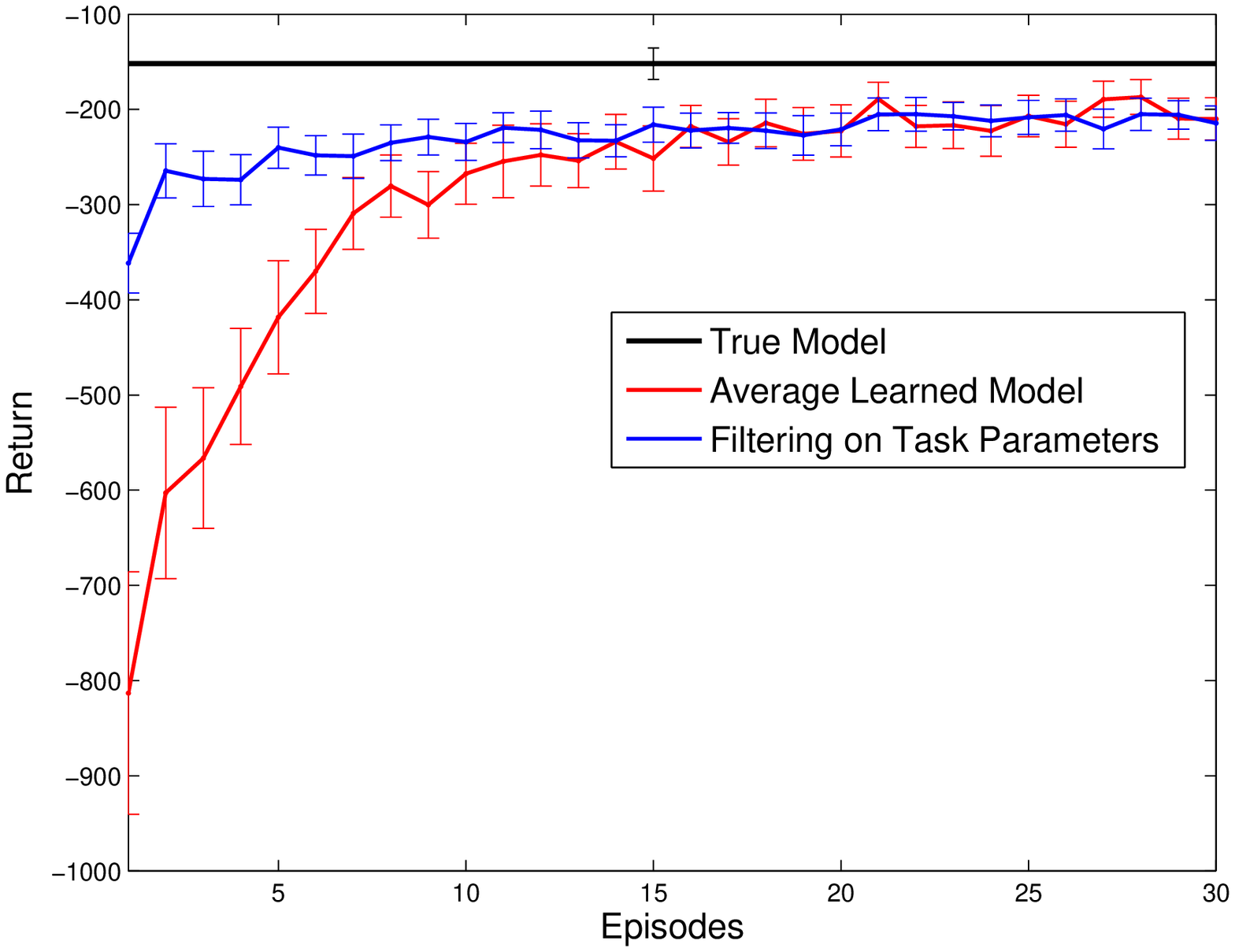}
%\vspace{-10pt}
}{
\caption{Acrobot performance. The HiP-MDP quickly approaches
  near-optimal performance.}
  \label{fig:acro_results}
}
\capbtabbox{
\small
\begin{tabular}{|p{.12in}|p{.47in}|p{.47in}|p{.47in}|p{.55in}|} \hline
\input{acro_exact_mse_output.txt}
\end{tabular}
}{
\caption{Mean-Squared Error on Acrobot (with 95\% confidence
  intervals)} 
  \label{tab:acrobot_exact_mse}
}
\end{floatrow}
\end{figure*}

Table~\ref{tab:acrobot_exact_mse} shows mean-squared prediction errors
using the same evaluation procedure as cartpole.  While the HiP-MDP
has slightly higher mean-squared errors on the angle predictions, the
IBP-GP approach does better overall and has lower mean-squared errors
on the angular velocity predictions.  Predicting angular velocities is
critical to planning in the acrobot task, as inaccurate predictions
will make the agent believe it can reach the swing-up position more
quickly than is physically possible.

In acrobat, a policy
learned with one mass setting will generally perform poorly on another.
%mass setting.  
For each of the 16 $( m_1 , m_2 )$ settings with $m_1
\in \{ .7 , .9 , 1.1 , 1.3 \}$ and $m_1 \in \{ .7 , .9 , 1.1 , 1.3
\}$, we ran 30 repeated trials in which we filtered the weight
parameters $w_k$ during an episode and updated the policy at the end
of each episode.  Even in this ``mostly observed'' setting, finding
the full Bayesian RL solution is PSPACE-complete~\cite{fern} and
offline approximation techniques are an active area of current
research~\cite{learning-to-plan}.  To avoid this complexity, 
we performed planning using Sarsa with dynamics
based on the mean weight parameters $w_k$. We first obtained
an initial value function using the prior mean weights,
and then interleaved model-based planning and reinforcement learning, 
using $5$ simulated episodes of planning for each episode of
interaction.

Results on acrobot are shown in figure~\ref{fig:acro_results}. 
We compare performance (averaged over weight settings)
to planning using the true model, and to planning using an average model
(learned using all the batch data at once) to obtain an initial value function. 
Both the average and HiP-MDP model reach performance
near, but not quite as high as, using the true model.
However, the HiP-MDP model is already near that performance by the
$2$nd episode.  Using the learned bases from the batches, as well as
learning the weights from the first episode, lets it quickly ``snap''
very near to the true dynamics.  By 5 episodes, our IBP-GP model has
reached near-optimal performance; the average model takes 15 episodes
to reach this level.

\section{Discussion and Related Work}

Bai et al. \cite{learning-to-plan} use a very similar hidden-parameter
setting treated as a POMDP to perform Bayesian planning; they assume
the model is given, whereas our task is to learn it.  The HiP-MDP is
similar to other POMDPs with fixed hidden states, for example, POMDPs
used for slot-filling dialogs \cite{sumpomdp}.  The key difference,
however, is that the objective of the HiP-MDP is not simply to gather
information (e.g., simply learn the transition function $T$); it is to
perform well on the task (e.g., drive a new car).  Transfer
learning---the goal of the HiP-MDP---has received much attention in
reinforcement learning.  Most directly related in spirit to our
approach are Hidden-Goal MDPs~\cite{fern} and hierarchical model-based
transfer~\cite{wilson-discrete}.  In both of these settings, the agent
must determine its current MDP from a discrete set of possible MDPs
rather than over continuous parameter settings.  More related from a
technical perspective are representation transfer approaches
\cite{Ferguson06,Taylor07,Ferrante08}, which typically learn a set of
basis functions sufficient for representation any value function
defined in a specific state space, or on transfer between two
different representations of the same task. By contrast, the HiP-MDP
focuses on modeling the dimensions of variation of a family of related
tasks.

The IBP-GP prior itself relates to a body of work on multiple-output
Gaussian processes (e.g. \cite{conv_multi_gp, teh_semi, gp_network}),
though most of these are focused on learning a convolution kernel to
model several related outputs on a single task, rather than
parameterizing several related tasks.  Gaussian process latent
variable models \cite{gplvm} have been used for dimensionality
reduction in dynamical systems.  As with other multi-output GP models,
however, GP-LVMs find a time-varying, low-dimensional representation
for a single system, while we characterize each instance in a set of
systems by a stationary, low-dimensional vector.  The focus of these
efforts has also been on modeling rather than control.

The extensive work on scaling inference in these Gaussian process
models \cite{conv_multi_gp, var_sparse_gp, var_dynamical_gp} provides
several avenues for relaxing some of the approximations that we made
in this work (while adding new ones).  In settings where one may not
have an initial batch of data from several instances, a fully Bayesian
treatment of the filter parameters $z_{kad}$ and the basis functions
$f_{kad}$ might allow the agent to more accurately navigate its
exploration-exploitation trade-offs.  Exploring which uncertainties
are important to model---and which are not---is an important question
for future work.  Other extensions within this particular model
include applying clustering or more sophisticated hierarchical methods
to group together basis functions $f_{kad}$ and thus share statistical
strength.  For example, one might expect that ``opposing actions,''
such as in cartpole, could be decomposed into similar basis functions.

\section{Conclusion}

Machine learning approaches for control often train agents to expect
repeated experiences the same domain.  However, a more accurate model
is that the agent will experience repeated domains that vary in
limited and specific ways.  In this setting, traditional planning
approaches, which typically rely on models, may fail due to the large
inter-instance variation.  By contrast, reinforcement learning
approaches, which typically assume that the dynamics of a new instance
are completely unknown, fail to leverage information from related
instances.

Our HiP-MDP model explicitly models this inter-instance variation,
providing a compromise between these two standard paradigms for
control.  Our HiP-MDP model will be useful when the family of domains
has a parametrization that is small relative to its model and the
objectives remains similar across domains.  Many
applications---such as handling similar objects, driving similar cars,
even managing similar network flows---fit this scenario, where batch
observational data from related tasks are easy to obtain in advance.
In such cases, being able to generalize dynamics from only a few
interactions with a new operating regime (using data from many prior
interactions with similar systems) is a key step in building
controllers that exhibit robust and reliable decision making while
gracefully adapting to new situations.

\bibliographystyle{unsrt}
\small
\bibliography{hip_mdp}
\end{document}